# Fast Fourier single-pixel imaging using binary illumination

**ZIBANG ZHANG,[1] XUEYING WANG,[1] AND JINGANG ZHONG[1,2,*]**

[1]*Department of Optoelectronic Engineering, Jinan University, Guangzhou 510632, China*
[2]*Guangdong Provincial Key Laboratory of Optical Fiber Sensing and Communications, Jinan University, Guangzhou 510632, China*
*\*Corresponding author: tzjg@jnu.edu.cn*

**Abstract:** Fourier single-pixel imaging (FSI) has proven capable of reconstructing high-quality two-dimensional and three-dimensional images. The utilization of the sparsity of natural images in Fourier domain allows high-resolution images to be reconstructed from far fewer measurements than effective image pixels. However, applying original FSI in digital micro-mirror device (DMD) based high-speed imaging system turns out to be challenging, because the original FSI uses grayscale Fourier basis patterns for illumination while DMDs generate grayscale patterns at a relatively low rate. DMDs are a binary device which can only generate a black-and-white pattern at each instance. In this paper, we adopt binary Fourier patterns for illumination to achieve DMD-based high-speed single-pixel imaging. Binary Fourier patterns are generated by upsampling and then applying error diffusion based dithering to the grayscale patterns. Experiments demonstrate the proposed technique able to achieve static imaging with high quality and dynamic imaging in real time. The proposed technique potentially allows high-quality and high-speed imaging over broad wavebands.

## 1. Introduction

Single-pixel imaging [1-24] is a novel computational imaging scheme, allowing an image to be captured by using a detector without spatial resolution. Different from conventional imaging that performs direct imaging with a single shot, single-pixel imaging reconstructs the image computationally with multiple measurements. In the scheme of single-pixel imaging, an object is under time-varying illumination and a single-pixel detector collects the intensity of the resultant light field which has interacted with the object. Contemporary single-pixel imaging architecture develops from ghost imaging [1-3] and has been adapted to other fields, such as terahertz imaging [4, 5], three-dimensional (3-D) imaging [6-11], and multispectral or hyperspectral imaging [12, 13]. Single-pixel detectors, compared with conventional pixelated detectors (such as CCD and CMOS), have a few advantages, such as 1) to manufacture single-pixel detectors that can work over broad wavebands is easier and more cost-effective than pixelated detectors; 2) single-pixel detectors can be flexibly placed [18]. Given such advantages, single-pixel imaging finds various applications such as microcopy [14-16], imaging through turbid media [17, 18], and remote sensing [19, 20]. It has attracted considerable attention in recent years.

Single-pixel imaging is able to reconstruct an object image by acquiring the spatial information of the object with measurements in a sequence. The object is illuminated by a sequence of time-varying illumination patterns. The intensity of the resultant light field is then collected by a single-pixel detector. The light field which has interacted with the object carries the spatial information of the object. Mathematically speaking, single-pixel imaging composes the object image by obtaining the inner products of the object under view with the illumination patterns. As the throughput of imaging system is limited by the only one available pixel, it leads to the fact that spatial information acquisition is at the expense of time. The spatial information acquisition time $t_A$ depends on both the number of measurements $M$ and the measurement rate $R$: $t_A = M/R$. To reduce the spatial information acquisition time, one can either increase the measurement for a greater $R$ or reduce the number of measurements for a smaller $M$. For the measurement rate, it is mainly determined by the illumination patterns generation rate. If the illumination patterns are generated by using a spatial light modulator (SLM), the rate will be determined by the utilized SLM. Currently, digital micro-mirror device (DMD) is a popularly used SLM, for it can generate high-contrast predefined patterns at a high speed. State-of-the-art DMDs can generate ~20,000 binary patterns per second. Using multiplexing techniques [21] is an effective approach to accelerate the imaging system, for they allow parallel measurement, equivalently increasing the measurement rate. However, multiplexing techniques are at the expense of additional devices. Therefore, the most straightforward way to maximize the pattern illumination rate is to make full use of the allowable rate of the DMD. For the number of measurements, it mainly depends on the number of effective reconstructed pixels. In terms of perfect reconstruction, the more effective pixels are, the more measurements will be needed. The utilization of generic compressive sensing (CS) algorithms [22] or the prior of object images enables to reduce measurements, but the CS algorithms generally consume relatively large computational time.

For single-pixel imaging, both quality and efficiency are desirable. Recently, a single-pixel imaging technique which can produce high-quality two-dimensional (2-D) images is proposed, termed Fourier single-pixel imaging (FSI) technique [18]. The technique has also been proven capable of achieving sub-millimetric depth accuracy in 3-D imaging [10]. FSI uses Fourier basis (sinusoidal intensity) patterns for illumination so as to acquire the Fourier spectrum of the object image. FSI takes the advantage that the Fourier basis patterns can be generated by a few means, for example, by using a spatial light modulator or by using interference. However, to apply FSI in a high-speed imaging system remains a challenge, because the technique does not lend itself well to DMDs which are a binary device. The original FSI uses grayscale patterns for illumination but DMDs generate grayscale patterns far slower than binary patterns. It is because DMDs generate each grayscale pattern by using temporal dithering. Each grayscale pattern is a temporal mean of several binary patterns. DMDs generate a grayscale pattern by sequentially switching multiple binary patterns. Temporal dithering is at the expense of time. Hadamard single-pixel imaging lends itself well to DMDs, because Hadamard basis patterns are naturally binary. Thus, real-time Hadamard single-pixel imaging is relatively easy to achieve and relative techniques have been reported [8, 12, 21, 24]. It is also essential to adapt the original FSI to a high-speed imaging technique. As FSI is on the base of Fourier analysis, it allows direct acquisition of the Fourier transform of an image. And more importantly, Fourier transform is a transformation that naturally exists. With the well-developed Fourier analysis theory, the relative techniques, and the prior of natural images' sparsity in the Fourier domain, FSI can find various applications in many fields. For example, our previously proposed 3-D Fourier single-pixel technique [10], which successfully combines 2-D FSI with Fourier fringe projection profilometry, achieves high-resolution 3-D reconstruction of millmetric accuracy by subsampling the Fourier spectrum with a very low compression rate and negligible computational time.

In this paper, we propose to use spatial dithering to binarize Fourier basis patterns for illumination. It allows us to take the advantage of high-speed binary patterns illumination ability of a DMD so as to speed up data acquisition. As such, fast high-quality static and real-time dynamic single-pixel imaging can be achieved.

## 2. Principles

The FSI uses Fourier basis patterns for illumination. The intensity of each pattern is sinusoidal. Therefore, each Fourier basis pattern, $P_\phi(x,y)$, is essentially a grayscale pattern and characterized by its spatial frequency pair $(f_x, f_y)$ and its initial phase $\phi$:

$$P_\phi(x,y) = a + b \cdot \cos(2\pi f_x x + 2\pi f_y y + \phi), \tag{1}$$

where $(x, y)$ is 2-D Cartesian coordinates, $a$ is the average intensity of the pattern, and $b$ the contrast. However, the original FSI does not lend itself well to DMD-based imaging systems, because Fourier basis patterns are of multiple gray scales while DMDs are a binary device. DMDs have millions of mirrors and each mirror has only two states ('ON' and 'OFF'). When a mirror is 'ON', it reflects the light toward the object to be illuminated; when the mirror is 'OFF', it reflects the light toward other direction. Each mirror can be individually controlled and switches between the two states. Therefore DMDs can generate various different binary (black-and-white) patterns, but can't generate grayscale patterns directly. As Figure 1(a) shows, to generate grayscale patterns, DMD-based projectors are designed to decompose each grayscale pattern into a sequence of binary patterns (also known as bitplanes). For example, a grayscale pattern with 256 gray levels (8-bit image) will be decomposed into 8 binary patterns (1-bit images). Then the resultant binary patterns are displayed on the DMD sequentially in a predefined amount of time. As the time for displaying each binary pattern is very short, the quick switching of binary patterns visually results in a grayscale pattern. Each pixel's intensity in the grayscale pattern is the temporally weighted (by the predefined amount of time) mean of the intensities of the 8 binary patterns at the same pixel. Such a strategy is termed temporal dithering which makes use of temporal average effect. It can be seen that DMDs achieve grayscale patterns generation at the expense of time (or temporal resolution). Even state-of-the art DMDs can only generate 8-bit grayscale patterns at a rate of ~250 Hz which is far lower than the rate (~20,000 Hz) for binary patterns. Therefore, the original FSI is difficult to be applied for high-speed imaging even by using a DMD.

In order to achieve high-speed Fourier basis patterns generation by using a DMD, we propose to generate binary Fourier basis patterns for illumination via spatial dithering. We use Floyd-Steinberg dithering algorithm [26] for the purpose of spatial dithering and the algorithm is based on the idea of error diffusion. Binarization inevitably causes quantization errors. The errors can be positive or negative. The idea of error diffusion is to spread (add) the residual quantization error of a pixel onto its neighboring pixels. If the quantization errors of a number of pixels have been negative, it becomes more likely that the quantization of next pixel is positive. As such, the quantization error in a local is close to zero on average, resulting in a visually grayscale pattern. In order to further eliminate the errors, we also propose to apply upsampling to the original grayscale pattern before dithering. The whole procedure of a binary Fourier basis pattern via the proposed spatial dithering strategy is shown in shown Figure 1(b).

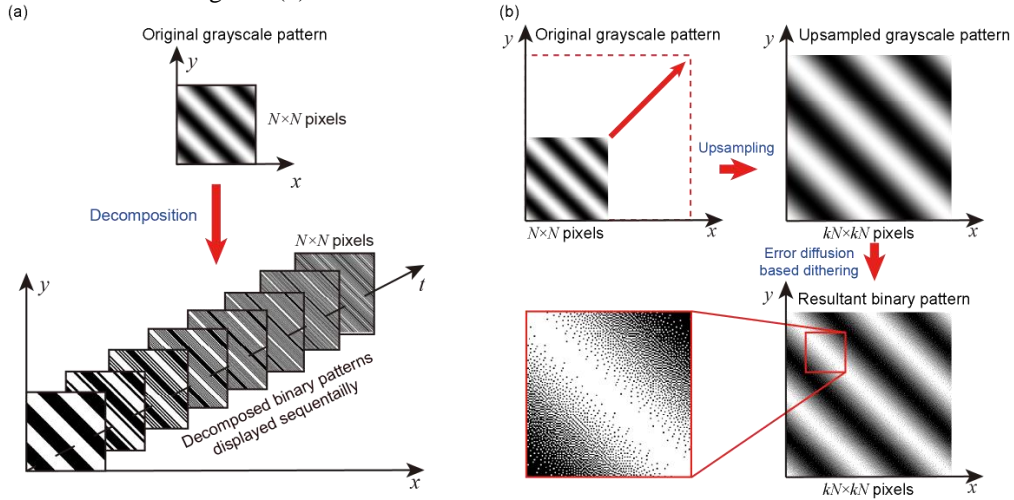

Fig. 1. Fourier basis patterns illumination by (a) temporal dithering and (b) spatial dithering.

To generate a binary Fourier basis pattern, we firstly upsample the original low-resolution ($N \times N$-pixel) grayscale pattern to be a high-resolution ($kN \times kN$-pixel) pattern. As such, each pixel in the original pattern will be represented by $k \times k$ pixels in the upsampled pattern, which is similar to $k \times k$-pixel binning [23]. We then apply an error diffusion based dithering algorithm to the upsampled pattern, with which a binary Fourier basis pattern is derived.

This strategy allows us to implement FSI by using binary patterns for illumination and to take the high-speed binary patterns generation ability given by a DMD. We use a sequence of $kN \times kN$-pixel binary patterns for illumination to reconstruct a $N \times N$-pixel image. In other words, each pixel in the reconstructed image is the spatial average of the reflectance over the corresponding $k \times k$ pixels on the object surface. Benefiting from the error diffusion, the mean error within these $k \times k$ pixels tends to zero. A larger $k$ would lead to smaller errors but lower effective resolution. To sum up, we use upsampling and error diffusion based dithering to exploit the spatial average effect in binary Fourier basis patterns generation, transferring the DMD's expense from time to space. This strategy allows us to shorten the spatial information acquisition time $t_A$ by increasing the measurement rate $R$.

In order to further speed up data acquisition, we employ three-step phase-shifting illumination to reduce the number of measurements. Three-step phase-shifting algorithm [27, 28] is algorithm widely in fringe analysis. Here we employ the three-step phase-shifting algorithm in Fourier coefficients acquisition. Fourier basis patterns with different spatial frequency pairs are generated according to Eq. (1). Each spatial frequency pair $(f_x, f_y)$ corresponds to three different initial phases ($\phi = 0$, $2\pi/3$, and $4\pi/3$ rad). As such, each complex-valued Fourier coefficient $\tilde{I}$ can be obtained by illuminating three patterns and using the three corresponding responses ($D_0$, $D_{2\pi/3}$, and $D_{4\pi/3}$) for calculation:

$$\tilde{I}(f_x, f_y) = \left[ 2D_0(f_x, f_y) - D_{2\pi/3}(f_x, f_y) - D_{4\pi/3}(f_x, f_y) \right] + \sqrt{3}\mathrm{j} \cdot \left[ D_{2\pi/3}(f_x, f_y) - D_{4\pi/3}(f_x, f_y) \right]. \quad (2)$$

As the conjugated symmetry of the Fourier spectrum of a natural image allows perfect reconstruction by half sampling the spectrum, it consumes $1.5N^2$ ($= N \times N \times 3/2$) measurements in total to fully sample an image by using the three-step phase-shifting algorithm.

### 3. Experiments

The proposed binary Fourier basis patterns generation and the three-step phase-shifting strategies allow to increase the measurement rate and to reduce the number of measurements, respectively. With the proposed strategies, real-time single-pixel imaging becomes possible in principle. We present experiments to verify the proposed strategies and demonstrate the real-time imaging ability in practice.

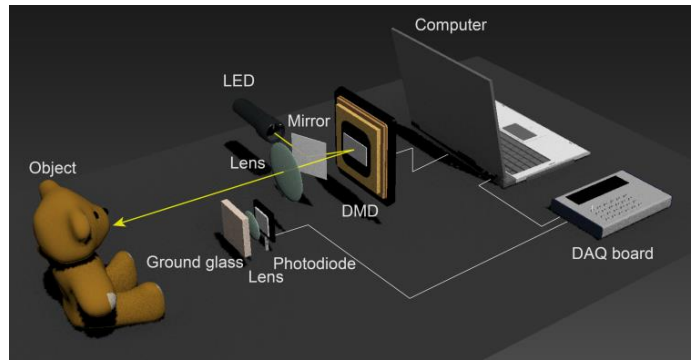

Fig. 2. Experimental set-up.

The experimental set-up is shown in Figure 2. It consists of an illumination system, a detection system, and an object. The illumination system consists of a Texas Instruments DLP Discovery 4100 development kit, a lens system, and a white LED. The DLP development kit is equipped with a 0.7-inch DMD which has 1024×768 micro mirrors. Each mirror is 13.6×13.6 μm$^2$ in size. The DLP development kit is equipped with an 8-GB RAM, allowing ~84000 512×512-pixel binary patterns to be stored on the board at one time. The LED continuously emits uniform light fields towards the mirror. With the mirror, light fields are reflected onto the DMD, then modulated by the DMD, and finally reflected out onto the object. Through the lens system, clear illumination patterns are formed on the object surface. Note that the lens system is simply removed from a commercial digital projector (TOSHIBA T-90). The illumination light fields are scattered by the object and the resulting scattered light fields are collected by the detection system. The detection system consists of a photodiode (HAMAMATSU S1227-1010BR) which is used as a single-pixel detector, a collecting lens, a piece of ground glass, a custom amplification circuit, a data acquisition board [National Instruments USB-6343 (BNC), and a computer. The maximum rate for analog input of the data acquisition board is 500,000 samples per second. The active area of the photodiode is 10×10 mm$^2$. The optimal rise time of the photodiode is 7 μs. In order to demonstrate that the proposed technique allows flexible placement of detector, we place the single-pixel detector behind a ground-glass diffuser so that the detector cannot 'see' the object directly. The thickness of the ground glass is 5 mm.

### 3.1 Static imaging

The static imaging experiments are to validate the proposed strategies and evaluate the achievable imaging quality of our technique. The object to be imaged is a toy and a piece of A4 paper with a printed enlarged 1951 USAF resolution test pattern. In order to reconstruct a 256×256-pixel image, we initially generate a complete set of grayscale Fourier basis patterns. The resolution of each pattern is 256×256 pixels. We then upsample the basis patterns by using bicubic interpolation algorithm with our predefined parameter $k = 2$ so that each upsampled pattern has 512×512 pixels. Floyd-Steinberg dithering algorithm is applied to all the patterns. For perfect reconstruction, all patterns are illuminated onto the object in order of spatial frequency from low to high. The detailed sampling strategy is described in Ref. [18].

Figure 3 shows the reconstructed images corresponding to different illumination rates ($R = 50$ Hz, 10,000 Hz, and 20,000 Hz). The spatial information acquisition time is 32.8 minutes, 9.83 seconds, and 4.91 seconds, respectively. Please note that the result corresponding to illumination rate of 50 Hz is done by using a 3 W LED as the illumination light source while the results corresponding to illumination rate of 10,000 Hz and 20,000 Hz are done by using a 60 W LED. We replace the light source because the light of the 3 W LED is too weak. Such weak luminance results in poor signal-to-noise ratio in high-speed measurement. It should be also noted that, as the illumination lens system is not customized, we can just couple a small portion of the light from the 60 W LED onto the DMD. As a result, we only obtain ~4-fold signal enhancement when using the 60 W LED.

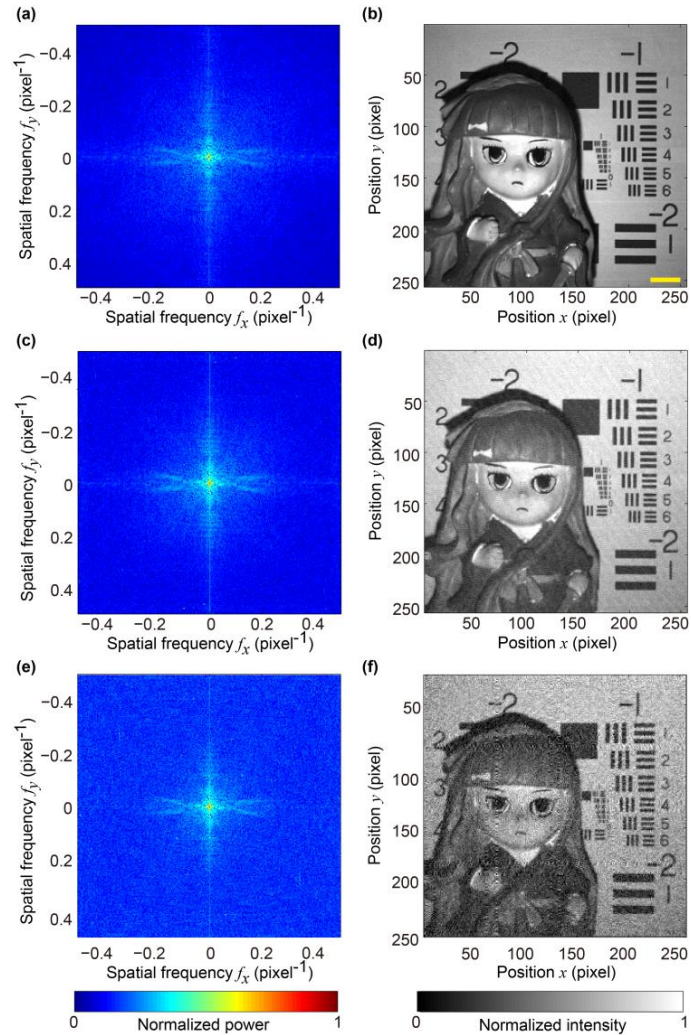

Fig. 3. Monochromatic static imaging. (a) Fourier spectrum acquired at illumination rate of 50 Hz and (b) corresponding reconstructed image; (c) and (d) at illumination rate of 10,000 Hz; (e) and (f) at illumination rate of 20,000 Hz. Note that we show the absolute value of the Fourier spectra on a logarithm scale to render them visible. No post processing has been applied to all reconstructed images. Scale bar = 2 cm.

The result for illumination rate of 50 Hz appears most clear, although the luminance is weak. The relatively longer acquisition time for each illumination patterns well evens out the random noise in the data acquisition. It should be emphasized that, although dithering causes quantization errors, noise is hardly observed in the reconstructed image even we use the parameter $k = 2$. In contrast, as the illumination rate increases, noise tends to be obvious in the reconstructed images. The quality of reconstruction degrades slightly when the illumination rate increases from 50 Hz to 10,000 Hz, but degrades dramatically when the illumination rate increases from 10,000 Hz to 20,000 Hz. We consider that there are two reasons and provide Figure 4 for further explanations.

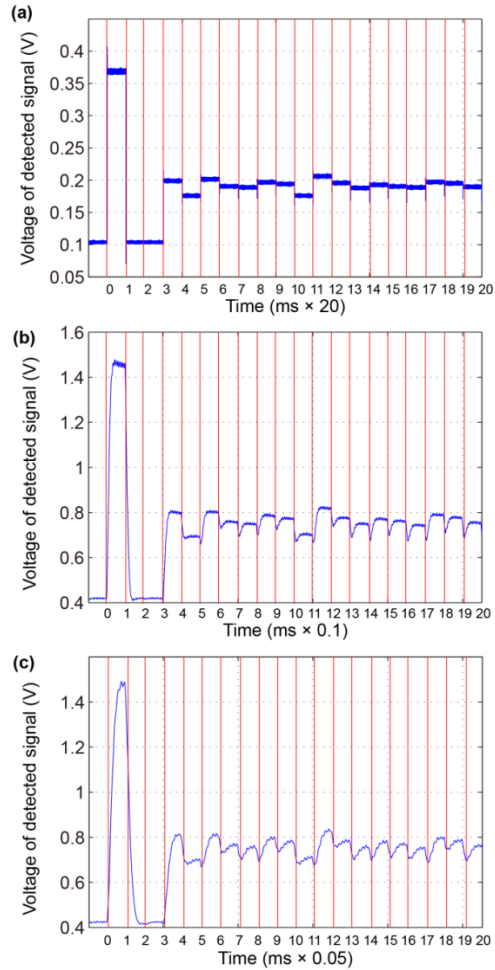

Fig. 4. Raw detected signals corresponding to the first 20 illumination patterns at the illumination rate of (a) 50 Hz, (b) 10,000 Hz, and (c) 20,000 Hz.

The figure shows the raw detected signals corresponding to the first 20 illumination patterns at the illumination rate of 50 Hz, 10,000 Hz, and 20,000 Hz, respectively. We consider the first reason is that random noise presents throughout the data acquisition process. Higher illumination rate results in shorter acquisition time for each illumination pattern and, consequently, each detection value is from fewer sampling points. Thus, the random noise is not well evened out with a relatively small number of sampling points, which results in low signal-to-noise ratio. For instance, when the illumination rate is 50 Hz, each detection value is the average of 10,000 sampling points; but when the illumination rate is 20,000 Hz, there will be only 25 sampling points remaining. The second reason is that response (bandwidth) of the detection system is not high enough for the illumination rate of 20,000 Hz. It can be observed from Figure 4 that the delay of the response tends to obvious, as the illumination rate increases. The delay is determined by the rise time of the utilized detector and the cut-off frequency of the amplification circuit for single-pixel detector. The delay tends to obvious when the rate is high, shortening the time of stationary state. Non-stationary sampling points contribute to noise. Although the reconstruction at 20,000 Hz appears relatively noisy, the object can be still clearly recognized. We believe that the first problem might be relieved by using a DAQ board with higher detection rate while the second problem by using a single-pixel detector with high-speed response and an optimized amplification circuit.

We also apply the proposed technique in true-color imaging. We use a color wheel filter for red, green, and blue illuminations. The filter is removed from a digital projector and placed between the LED and the mirror. In this experiment, we move the toys closer to the lens system and set the focus of the lens system on the toys. As such, the toys would appear clear while the background would appear blurred in the reconstructed image. We also move the photodiode, the ground glass, and the collecting lens above the DMD so as to avoid generating apparent shadows in the reconstructed images. We derive three images for red, green, and blue channels, respectively. Each image is fully sampled and derived at the illumination rate of 50 Hz, consuming 32.8 minutes. The 3 W LED is used as the illumination light source. The acquisition of all three images consumes ~98 minutes. The derived images and the final true-color reconstruction are shown in Figure 5.

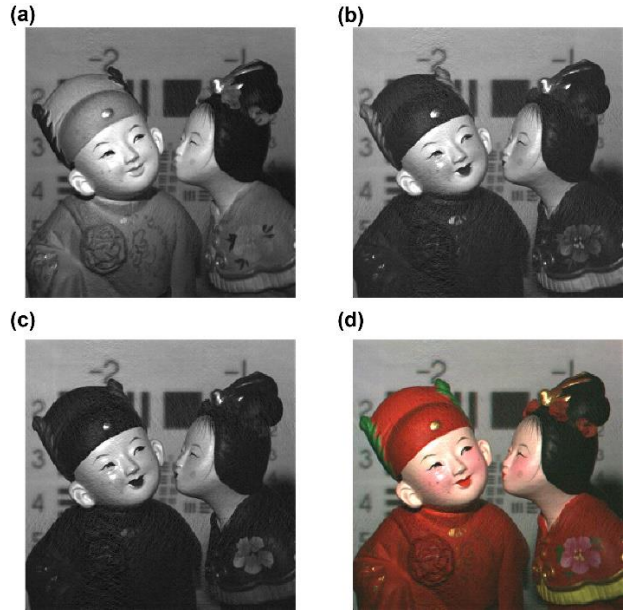

Fig. 5. True-color static imaging. (a) red channel image, (b) green channel image, (c) blue channel image, and (d) combined true-color image. No post processing has been applied to the reconstructed images.

### 3.2 Dynamic imaging

We finally apply the proposed technique in real-time imaging, capturing a dynamic scene. The illumination rate of the DMD is set to be 10,000 Hz, because the response of the utilized photodiode is too low for the rate of 20,000 Hz. The illumination patterns initially have $128 \times 128$ effective pixels and are upsampled to be $512 \times 512$ pixels (that is, $k = 4$). With the prior that most spatial information of natural images is concentrated in the low-frequency range in the Fourier domain, we only sample 333 coefficients in the low-frequency range along a spiral path for each image (the detailed sampling strategy is described in Ref. [18]). The rate of compression is ~4% ($= 333 \times 2/128^2$). As such, we can capture ~10 images per second. We capture 258 images and reconstruct a ~26-second video. Figure 6 shows 9 among all images reconstructed. We consider that if given a detection system which has higher response for the illumination rate of 20,000 Hz, we would have a better result at the same frame rate.

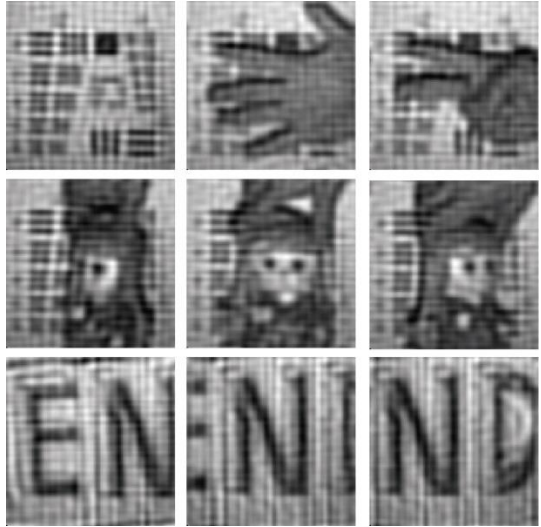

Fig. 6. Real-time dynamic imaging reconstruction. No post processing has been applied to the reconstructed images.

## 4. Discussion

We propose to use an error diffusion based dithering algorithm to generate binary Fourier basis patterns for illumination. This strategy has the advantages of 1) the illumination rate is remarkably increased, in comparison to grayscale Fourier basis patterns generation by using a DMD; 2) binary Fourier basis patterns are potentially more suitable for single-pixel microscopy. It is because illumination patterns used in microcopy are commonly of much higher spatial binary spatial frequency than illumination patterns used in photography. Due to the limited bandwidth of the optical transfer function, the binary Fourier basis patterns would naturally tend to sinusoidal through an objective lens, although the patterns generated by the DMD are initially binary. In this case, the proposed binary FSI might outperform single-pixel techniques that initially use binary patterns illumination; 3) binary patterns are free from gamma distortion (nonlinearity of gray levels); 4) each pixel of the binary patterns takes only 1 bit to store on the DMD's RAM, which reduces 87.5% storage space. It allows more patterns to be stored in the RAM at one time. However, the strategy has also disadvantages of 1) dithering causes quantization errors. Even the utilization of error diffusion based dithering cannot perfectly eliminate the errors. We experimentally demonstrated the errors are unnoticeable in the final reconstruction even we use the parameter $k = 2$ in the conversion of a 256-level grayscale image to a binary image; 2) high-speed binary Fourier basis patterns generation is at the expense of spatial resolution. The larger $k$ would lead to the lower spatial resolution. We recommend using $k = 2$, as it is experimentally demonstrated able to give highest resolution with unnoticeable noise level in reconstruction.

The use of the three-step phase-shifting algorithm brings an advantage of 25% fewer measurements than the previously the four-step phase-shifting algorithm used in our previously proposed technique and other single-pixel imaging techniques where differential measurement is employed. Although two-step phase-shifting ($\pi$-shift) strategy, which is a direct method of measurement, allows to further reduce the number measurements in principle, such an algorithm is not robust against noise, because noise always exists in practice.

The achievable highest acquisition rate depends on the rate of the utilized DMD and the response of the utilized detector. The single-pixel detector we use in our experiments is a photodiode. The photodiode has a relatively large active area and therefore has high sensitivity but low-speed response. The quality of the results for the high-speed imaging in our experiments is mainly restricted by the response speed of the detection. We consider that the use of a high-speed response detector (such as, PIN photodiode or photomultiplier tube) can further improve the quality of reconstructions.

FSI obtains the desired spatial information by performing Fourier basis scanning so as to acquire the Fourier spectrum of the object image. The sparsity of natural images results in only a relatively small number of Fourier coefficients being large in the Fourier space. The use of advanced sampling strategies [29, 30] in Fourier coefficients acquisition might allow further reduction of measurements.

FSI is computationally effective, because in the image reconstruction process it only employs inverse Fourier transform and the transform has fast algorithms. The computational time $t_\mathrm{C}$ is negligible in comparison with the spatial information acquisition time $t_\mathrm{A}$.

The proposed technique allows real-time Fourier spectrum acquisition by using a single-pixel detector. The proposed technique can combine with the 3-D Fourier single-pixel imaging technique to achieve real-time 3-D single-pixel imaging.

## 5. Conclusion

We adopt spatially dithering Fourier basis patterns for illumination to accelerate the imaging speed in a DMD-based single-pixel imaging system. The proposed technique is experimentally demonstrated capable of producing high-quality images in static imaging and capturing dynamic evens in real-time imaging. The technique might open many perspectives in applications where invisible wavebands are required.

**Funding.** National Natural Science Foundation of China (NSFC) (61475064).

**Acknowledgments**. The authors thank Dr. Shiping Li for her help with the experimental equipment preparation, Bingyong Huang for his help with the custom circuit design, and Qinqiu Fang for linguistic assistance.